\documentclass[letterpaper, 10 pt, journal, twoside]{IEEEtran}
\usepackage{amsmath} 
\usepackage{amssymb}
\usepackage{siunitx}
\usepackage{caption}
\usepackage{graphicx}
\usepackage{float}
\usepackage{cite}
\usepackage{color}

\ifCLASSINFOpdf
\else
\fi
\newcommand{\ignore}[1]{}

\newcommand{\mc}[1]{\mathcal{#1}}  

\newcommand{\bbm}{\begin{bmatrix}}
\newcommand{\ebm}{\end{bmatrix}}
\newcommand{\bma}[1]{\left[\begin{array}{#1}}
\newcommand{\ema}{\end{array}\right]}

\DeclareMathAlphabet{\mbf}{OT1}{ptm}{b}{n}
\newcommand{\mbs}[1]{{\boldsymbol{#1}}}

\newcommand{\mbsdot}[1]{{\dot{\boldsymbol{#1}}}}

\newcommand{\mbshat}[1]{{\hat{\boldsymbol{#1}}}}

\newcommand{\mbsdel}[1]{{\delta {\boldsymbol{#1}}}}

\newcommand{\mbfdot}[1]{{\dot{\mbf{#1}}}}
\newcommand{\mbfbar}[1]{{\bar{\mbf{#1}}}}
\newcommand{\mbfhat}[1]{{\hat{\mbf{#1}}}}

\newcommand{\mbfdel}[1]{{\delta{\mbf{#1}}}}

\newcommand{\mbfch}[1]{{\check{\mbf{#1}}}}
\newcommand{\mbsch}[1]{{\check{\boldsymbol{#1}}}}
\newcommand{\mbfcheck}[1]{{\check{\mbf{#1}}}}

\def\fdotb{{\raisebox{-0.6ex}{ \kern0.2ex\raisebox{0.8ex}{\tiny $\hspace*{-1ex}\circ$}}}}
\def\fddotb{{\raisebox{-0.6ex}{ \kern0.2ex\raisebox{0.8ex}{\tiny $\hspace*{-1ex}\circ\circ$}}}}

\newcommand{\f}{\frac}

\newcommand{\trans}{{\ensuremath{\mathsf{T}}}} 
\newcommand{\utimes}{ {\raisebox{-0.6ex}{ \kern-1.0ex\raisebox{0.6ex}{ \small $\mathsf{v}$}}} } %
\newcommand{\beq}{\begin{equation}}
\newcommand{\eeq}{\end{equation}}
\newcommand{\bdis}{\begin{displaymath}}
\newcommand{\edis}{\end{displaymath}}
\newcommand{\beqarray}{\begin{eqnarray}}
\newcommand{\eeqarray}{\end{eqnarray}}
\newcommand{\beqarraynn}{\begin{eqnarray*}}
\newcommand{\eeqarraynn}{\end{eqnarray*}}

\renewcommand{\theenumii}{\arabic{enumii}}

\makeatletter
\renewcommand{\p@enumii}{\theenumi.}
\makeatother

\newtheorem{theorem}{Theorem}

\begin{document}

\newpage
%
%
%
%
%
%
%
\def \myJournal {IEEE Robotics and Automation Letters}
\def \myDoi {10.1109/LRA.2020.3005132}
\def \myPaperSiteName {IEEE Xplore}
\def \myPaperSiteLink {https://ieeexplore.ieee.org/document/9126191}
\def \myYear {2024}

\def \myPaperCitation{N. van der Laan, M. Cohen, J. Arsenault, and J. R. Forbes, ``The Invariant Rauch-Tung-Striebel Smoother,'' in \textit{IEEE Robotics and Automation Letters}, vol. 5, no. 4, pp. 5067-5074, June 2020.}


\begin{figure*}[t]

\thispagestyle{empty}
\begin{center}
\begin{minipage}{6in}
\centering
This paper has been accepted for publication in \emph{\myJournal}. 
\vspace{1em}

This is the author's version of an article that has, or will be, published in this journal or conference. Changes were, or will be, made to this version by the publisher prior to publication.
\vspace{2em}

\begin{tabular}{rl}
DOI: & \myDoi\\
\myPaperSiteName: & \texttt{\myPaperSiteLink}
\end{tabular}

\vspace{2em}
Please cite this paper as:

\myPaperCitation

\vspace{15cm}
\copyright \myYear \hspace{4pt}IEEE. Personal use of this material is permitted. Permission from IEEE must be obtained for all other uses, in any current or future media, including reprinting/republishing this material for advertising or promotional purposes, creating new collective works, for resale or redistribution to servers or lists, or reuse of any copyrighted component of this work in other works.

\end{minipage}
\end{center}
\end{figure*}
\newpage
\clearpage
\pagenumbering{arabic} 
%
\title{The Invariant Rauch-Tung-Striebel Smoother}
%
%
%

\author{Niels van der Laan$^{1}$, Mitchell Cohen$^{2}$, Jonathan Arsenault$^{3}$ and James Richard Forbes$^{4}$
\thanks{Manuscript received: February 24th, 2020; Revised May 13th, 2020; Accepted June 8th, 2020.}
\thanks{This paper was recommended for publication by Editor Sven Behnke upon evaluation of the Associate Editor and Reviewers' comments.
This work was supported by the Group for Research in Decision Analysis (GERAD) and the National Science and Engineering Research Council (NSERC) of Canada.}
\thanks{$^{1}$Research Trainee, Group for Research in Decision Analysis (GERAD), Montreal, QC, H3T 1J4, Canada. Email: n.vanderlaan-2@student.tudelft.nl}%
\thanks{$^{2}$M.Eng. Candidate, Department of Mechanical Engineering,
McGill University, Montreal, QC, H3A 0C3, Canada. Email: mitchell.cohen3@mail.mcgill.ca}
\thanks{$^{3}$Ph.D. Candidate, Department of Mechanical Engineering,
McGill University, Montreal, QC, H3A 0C3, Canada. Email: jonathan.arsenault@mail.mcgill.ca}%
\thanks{$^{4}$Associate Professor, Department of Mechanical Engineering, McGill University, Montreal, QC, H3A 0C3, Canada. Email: james.richard.forbes@mcgill.ca}%
\thanks{Digital Object Identifier (DOI): see top of this page.}
}
%
%

\markboth{IEEE Robotics and Automation Letters. Preprint Version. Accepted June, 2020}
{van der Laan \MakeLowercase{\textit{et al.}}: The Invariant Rauch-Tung-Striebel Smoother} 

%



\maketitle

\begin{abstract}
This paper presents an invariant Rauch-Tung-Striebel (IRTS) smoother applicable to systems with states that are an element of a matrix Lie group. In particular, the extended Rauch-Tung-Striebel (RTS) smoother is adapted to work within a matrix Lie group framework. The main advantage of the invariant RTS (IRTS) smoother is that the linearization of the process and measurement models is independent of the state estimate resulting in state-estimate-independent Jacobians when certain technical requirements are met. A sample problem is considered that involves estimation of the three dimensional pose of a rigid body on $SE(3)$, along with sensor biases. The multiplicative RTS (MRTS) smoother is also reviewed and is used as a direct comparison to the proposed IRTS smoother using experimental data. Both smoothing methods are also compared to invariant and multiplicative versions of the Gauss-Newton approach to solving the batch state estimation problem.
\end{abstract}


%
\IEEEpeerreviewmaketitle

\section{Introduction}
%
%
%
%
\IEEEPARstart{T}he need to estimate states using incomplete and noisy data arises in many engineering problems. In robotics applications, where states of interest are often elements of a matrix Lie group, a popular state estimator is the extended Kalman filter (EKF), or rather the multiplicative extended Kalman filter (MEKF), a variant of the EKF \cite{MyPaper:Lefferts}. The EKF is an approximation to the Bayes filter that uses linearization to compute a state estimate \cite[Sec.~4.2]{MyPaper:barfoot}. The prediction and correction steps of the EKF use process and measurement model Jacobians, respectively, that are evaluated at the most recent state estimate. If the most recent state estimate is poor, the Jacobians are inaccurate.

The invariant EKF (IEKF) is a state estimator that is specific to systems with states that are an element of a matrix Lie group. It was shown in \cite{MyPaper:Barrau,MyPaper:Barrau_2} that when the process model is group affine, the measurements are left invariant (right invariant), and a left-invariant error (right-invariant error) is used, the IEKF as interpreted as an observer possesses asymptotic convergence properties. Moreover, the Jacobians associated with the IEKF, when treated as an observer, are state-estimate independent. As such, the IEKF has enhanced performance properties when compared to the EKF or MEKF \cite{MyPaper:Barrau, MyPaper:Barrau_2, Arsenault2019}. 

Unlike the Kalman filter and its nonlinear variants, the Rauch-Tung-Striebel (RTS) smoother uses all available measurements in a batch estimation framework to compute state estimates using a forward pass followed by a backward pass. In \cite{MyPaper:bourmaud} an RTS smoother for systems with states that are an element of a matrix Lie group is presented, but the invariant framework is not leveraged. Batch estimation in an invariant framework is considered in \cite{MyPaper:Chauchat2018}, with additional details and extensions given in \cite{Chauchat2020thesis}, leading to an invariant Gauss-Newton (IGN) algorithm. Invariant sliding window filtering is also considered in \cite{Walsh2019} in the context of attitude and bias estimation. Applying the invariant framework to the simultaneous localization and mapping problem in a batch framework is considered  in \cite{Arsenault2019}. The advantage of the invariant framework is the fact that the Jacobians are state-estimate independent. Motivated by this fact, this paper considers the derivation of an invariant RTS (IRTS) smoother for systems with states that are elements of a matrix Lie group. Like the IEKF and IGN, the IRTS smoother has state-independent Jacobians when certain technical requirements are met. 




To assess the performance of the IRTS smoother relative to a multiplicative RTS (MRTS) smoother, IGN, as well as multiplicative GN (MGN), a state estimation problem is considered that involves estimating the position and attitude of a rigid body. Two interoceptive sensor measurements, as well as two exteroceptive measurements, are available. Angular velocity and translational velocity measurements are the interoceptive measurements, while both left-invariant and right-invariant measurements from a GPS receiver and a stereo camera are the exteroceptive measurements. In addition, biases on the angular and translational velocity sensors are also estimated. Due to the inclusion of sensor biases, the state estimation problem does not satisfy the exact requirements of the invariant framework, that being a group-affine process model. However, leveraging the invariant framework still leads to Jacobians that depend less on the state estimates as compared to the multiplicative framework \cite{Arsenault2019}. The IRTS smoother, MRTS smoother, IGN, and MGN are tested and compared using the Starry Night dataset \cite{starryNightDataset}. The IRTS smoother outperforms MRTS smoothing, IGN and MGN when initial errors are large. The enhanced performance of the IRTS smoother stems from the fact that the Jacobians associated with the IRTS smoother are less state-estimate-dependent than those used in MRTS smoothing, IGN and MGN, and during the forward-backward passes of the IRTS smoother the Jacobians are updated at each step. 





The remainder of this paper is organized as follows. Notation and preliminaries are presented in Section~\ref{section:Preliminaries}. In Section~\ref{section:RTS} the standard RTS smoother is reviewed. An extended RTS smoother applicable to systems with states that are an element of matrix Lie group, the MRTS smoother, is presented in Section~\ref{section:MRTS}. Finally, the invariant framework is brought to bear on the RTS smoothing problem in Section~\ref{section:IRTS}. Section~\ref{section:Problem} formulates the $SE(3)$ with bias state estimation problem, while section~\ref{section:experimental_results} presents experimental results comparing the IRTS and MRTS smoothers, as well as IGN and MGN. The paper is drawn to a close in Section~\ref{section:Conclusions}. This paper's contributions are the IRTS smoother derivation in Section~\ref{section:IRTS} and its evaluation and comparison to three other batch estimation methods using experimental data in Section~\ref{section:experimental_results}. 

%

\section{Preliminaries}
\label{section:Preliminaries}

\subsection{Matrix Lie Groups}

Consider the matrix Lie group $\mathcal{G}$, which is composed of $n \times n$ matrices with $m$ degrees of freedom, that is closed under matrix multiplication \cite{MyPaper:Hall}. The matrix Lie algebra associated with $\mathcal{G}$, denoted by $\mathfrak{g}$, is the tangent space of $\mathcal{G}$ at the identity element $\mbf{1}$, denoted as $T_{\mbf{1}}\mathcal{G}$. The matrix Lie algebra can be mapped to the matrix Lie group using the exponential map, $\exp\left(\cdot\right)$ : $\mathfrak{g}\rightarrow\mathcal{G}$. The inverse map uses the matrix natural logarithm, $\log\left(\cdot\right)$ : $\mathcal{G}\rightarrow\mathfrak{g}$. The linear operator $\left(\cdot\right)^\wedge$ : $\mathbb{R}^m\rightarrow\mathfrak{g}$ maps an $m$ dimension column matrix to the matrix Lie algebra. The inverse map is $\left(\cdot\right)^\vee$ : $\mathfrak{g}\rightarrow\mathbb{R}^m$. Let $\textrm{Ad}(\cdot)$ denote the matrix representation of the adjoint operator, where $(\textrm{Ad}(\mbf{X})\mbs{\xi})^\wedge = \mbf{X}\mbs{\xi}^\wedge\mbf{X}^{-1}$ for $\mbf{X} \in \mc{G}$ and $\mbs{\xi} \in \mathbb{R}^m$. 


\textit{Definition 1 (Group Affine \cite{MyPaper:Barrau}):} The function $\mbf{F}\left(\mbf{X},\mbf{u}\right)$ is said to be group affine if for $\mbf{X}_1$, $\mbf{X}_2 \in \mathcal{G}$ it satisfies 
\beq \mbf{F}\left(\mbf{X}_1\mbf{X}_2,\mbf{u}\right)=\mbf{X}_1\mbf{F}\left(\mbf{X}_2,\mbf{u}\right)+\mbf{F}\left(\mbf{X}_1,\mbf{u}\right)\mbf{X}_2-\mbf{X}_1\mbf{F}\left(\mbf{1},\mbf{u}\right)\mbf{X}_2 . 
\label{eq:group_affine}
\eeq

\textit{Definition 2 (Left- and Right-Invariant Error \cite{MyPaper:Barrau}):} Consider $\mbf{X},\hat{\mbf{X}} \in\mathcal{G}$. The left-invariant error between $\mbf{X}$ and $\mbfhat{X}$ is given by $\delta\mbf{X}^\mathrm{L}=\mbf{X}^{-1}\hat{\mbf{X}}$ and the right-invariant error between $\mbf{X}$ and $\mbfhat{X}$ is $\delta\mbf{X}^\mathrm{R}=\hat{\mbf{X}}\mbf{X}^{-1}$.

\section{RTS Smoothing}
\label{section:RTS}

Consider the linear system 
\begin{align}
\mbfdot{x}(t) & = \mbf{A}\mbf{x}(t)+\mbf{B}\mbf{u}(t)+\mbf{L}\mbf{w}(t),\\
\mbf{y}_k  & = \mbf{H}_k\mbf{x}_k+\mbf{M}_k\mbf{v}_k \label{eq:RTS_DT_meas},
\end{align}
where $k$ denotes the time steps such that $\mbf{x}_k=\mbf{x}(t_k)$, $\mbf{x}(t)\in\mathbb{R}^m$ is the state, $\mbf{u}(t)$ is an interoceptive measurement, $\mbf{y}_k$ is the exteroceptive measurement, $\mbf{w}(t)$ is process noise, and $\mbf{v}_k$ is measurement noise. Unless required for clarity, the argument $(t)$ will be omitted for brevity. The discrete-time process model can be found in several ways \cite{MyPaper:Farrell} resulting in the discrete-time process model
\beq \label{eq:RTS_DT_process} \mbf{x}_k = \mbf{A}_{k-1}\mbf{x}_{k-1}+\mbf{B}_{k-1}\mbf{u}_{k-1}+\mbf{L}_{k-1}\mbf{w}_{k-1}. \eeq
The RTS smoother consists of a forward pass, equivalent to the Kalman filter, that goes forward in time and a backward smoothing pass that goes backward in time. 

The predicted state estimate and covariance are found using \cite[Sec.~8.2]{MyPaper:sarkka_2013}
\begin{align}
\check{\mbf{x}}_{f,k} & = \mbf{A}_{k-1}\hat{\mbf{x}}_{f,k-1}+\mbf{B}_{k-1} \mbf{u}_{k-1}, \\
\check{\mbf{P}}_{f,k} & = \mbf{A}_{k-1}\hat{\mbf{P}}_{f,k-1}\mbf{A}_{k-1}^\trans+\mbf{L}_{k-1}\mbf{Q}_{k-1}\mbf{L}_{k-1}^\trans, \label{eq:RTS_cov_check}
\end{align}
where $\mbf{w}_k \sim \mathcal{N}\left(\mbf{0},\mbf{Q}_{k}\right)$, $\check{\mbf{x}}_{f,k}$ is the \textit{a priori} forward state estimate, $\hat{\mbf{x}}_{f,k}$ is the \textit{a posteriori} forward state estimate, $\check{\mbf{P}}_{f,k}$ is the \textit{a priori} forward covariance estimate, $\hat{\mbf{P}}_{f,k}$ is the \textit{a posteriori} forward covariance estimate, and $k=0,\ldots,N$. The Kalman gain $\mbf{K}_{f,k}$ is computed using \cite[Sec.~8.2]{MyPaper:sarkka_2013}
\beq
\mbf{K}_{f,k} = \check{\mbf{P}}_{f,k}\mbf{H}_{k}^\trans\left(\mbf{H}_{k}\check{\mbf{P}}_{f,k}\mbf{H}_{k}^\trans+\mbf{M}_{k}\mbf{R}_k\mbf{M}_{k}^\trans\right)^{-1}, \label{eq:Gain_forward} 
\eeq
where $\mbf{v}_k \sim \mathcal{N}\left(\mbf{0},\mbf{R}_{k}\right)$.
The predicted estimates are then corrected using \cite[Sec.~9.4]{MyPaper:Simon}
\begin{align}
	\hat{\mbf{x}}_{f,k} & = \check{\mbf{x}}_{f,k}+\mbf{K}_{f,k}\left(\mbf{y}_k-\mbf{H}_{k}\check{\mbf{x}}_{f,k}\right), \label{eq:forward_correction} \\
	\hat{\mbf{P}}_{f,k} & = \left(\mbf{1}-\mbf{K}_{f,k}\mbf{H}_{k}\right)\check{\mbf{P}}_{f,k}\left(\mbf{1}-\mbf{K}_{f,k}\mbf{H}_{k}\right)^\trans \nonumber \\
	& +\mbf{K}_{f,k}\mbf{M}_{k}\mbf{R}_k\mbf{M}_{k}^\trans\mbf{K}_{f,k}^\trans. \label{eq:forward_cov_hat} 
\end{align}

After the forward pass is complete, the backward smoothing pass is initialized  using $\hat{\mbf{x}}_{f,N}$ and $\hat{\mbf{P}}_{f,N}$. The smoothing equations \cite[Sec.~8.2]{MyPaper:sarkka_2013}
\begin{align}
\mbf{K}_{s,k} & = \hat{\mbf{P}}_{f,k}\mbf{A}_k^\trans\check{\mbf{P}}_{f,k+1}^{-1}, \label{eq:RTS_smoother_gain} \\
\hat{\mbf{x}}_{s,k} & = \hat{\mbf{x}}_{f,k}+\mbf{K}_{s,k}\left(\hat{\mbf{x}}_{s,k+1}-\check{\mbf{x}}_{f,k+1}\right), \\
\hat{\mbf{P}}_{s,k} & = \hat{\mbf{P}}_{f,k}-\mbf{K}_{s,k}\left(\check{\mbf{P}}_{f,k+1}-\hat{\mbf{P}}_{s,k+1}\right)\mbf{K}_{s,k}^\trans, \label{eq:RTS_smoother_cov} 
\end{align}
are then used for $k=N-1,\ldots,0$, where $\mbf{K}_{s,k}$ is the smoother gain, $\hat{\mbf{x}}_{s,k}$ is the smoother estimate, and $\hat{\mbf{P}}_{s,k}$ is the smoother covariance.

\section{Multiplicative RTS Smoothing}
\label{section:MRTS}

The extended RTS smoother is used for nonlinear systems and has many variations, an example of which is discussed in \cite{MyPaper:sarkka_2013}. Consider the nonlinear continuous-time process and discrete-time measurement models
\begin{align}
\mbfdot{X}(t) & = \mbf{F}\left(\mbf{X}(t),\mbf{x}(t),\mbf{u}(t),\mbf{w}(t)\right), \label{eq:MRTS_CT_process1} \\
\mbfdot{x}(t) & = \mbf{f}\left(\mbf{X}(t),\mbf{x}(t),\mbf{u}(t),\mbf{w}(t)\right), \label{eq:MRTS_CT_process} \\
\mbf{y}_k & = \mbf{g}_k\left(\mbf{X}_k,\mbf{x}_k,\mbf{v}_k\right), \label{eq:MRTS_CT_meas}
\end{align}
where $\mbf{X} \in \mathcal{G}$ is an element of a matrix Lie group while $\mbf{x} \in \mathbb{R}^{n_x}$, and both compose the state. For a process model of the form given by \eqref{eq:MRTS_CT_process1} and \eqref{eq:MRTS_CT_process}, the MRTS smoother can be used for state estimation. A similar multiplicative approach to smoothing is taken in \cite{MyPaper:Psiaki, MyPaper:bourmaud}. The MRTS smoother can be viewed as an extension of the extended RTS smoother in the same way the multiplicative extended Kalman Filter (MEKF) is an extension of the standard EKF. Owing to the popularity of the EKF, and the use of the MEKF as the benchmark when assessing the benefits of the IEKF, the MRTS smoother will be compared to the proposed IRTS smoother derived in Section~\ref{section:IRTS}.

As is done in many Kalman filter variants, the MRTS smoother requires linearization of the  process and measurement models. Combining and linearizing \eqref{eq:MRTS_CT_process1} and \eqref{eq:MRTS_CT_process}, and also linearizing \eqref{eq:MRTS_CT_meas}, results in
\begin{align}
\delta\mbfdot{x} & = \mbf{A}\mbfdel{x}+\mbf{L}\delta\mbf{w} \label{MRTS_CT_process_linear},\\
\delta\mbf{y}_k & = \mbf{H}_k\mbfdel{x}_k+\mbf{M}_k\delta\mbf{v}_k,
\end{align}
where $\delta \mbf{w} = \mbfbar{w} + \mbf{w}$ and $\delta \mbf{v}_k = \mbfbar{v}_k + \mbf{v}_k$, $\mbf{A}$ and $\mbf{L}$ are the process model Jacobians, and $\mbf{H}_k$ and $\mbf{M}_k$ are the measurement model Jacobians. The error definitions used to perform the linearization are \cite{MyPaper:Aucoin}
\beq
\mbfdel{x} = \begin{bmatrix}
\log(\mbf{X}^{-1} \mbfhat{X})^\vee \\ 
\mbfhat{x}-\mbf{x}
\end{bmatrix}.
\eeq
The error in the state defined on $\mc{G}$ is multiplicative because $\mc{G}$ is not closed under addition. Also, this is just one error definition; alternative definitions can be used as well.



The forward filter is the known MEKF \cite{MyPaper:Lefferts,MyPaper:Psiaki}. The prediction step is performed by integrating \eqref{eq:MRTS_CT_process1} and \eqref{eq:MRTS_CT_process} from time $t_{k-1}$ to time $t_k$ using the expected value of the process noise, which is zero. The predicted covariance is given by \eqref{eq:RTS_cov_check} and the Kalman gain is given by \eqref{eq:Gain_forward}. The state estimate correction is given by
\begin{align}
\begin{bmatrix}
\mbsdel{\chi}_{1} \\ 
\mbsdel{\chi}_{2}
\end{bmatrix} & = \mbf{K}_{f,k}\mbf{z}_k, \\
\mbfhat{X}_{f,k} & = \check{\mbf{X}}_{f,k}\exp\left(-{\mbsdel{\chi}_{1}}^\wedge\right), \\
\mbfhat{x}_{f,k} & = \mbfch{x}_{f,k}-\mbsdel{\chi}_{2},
\end{align}
where $\mbf{z}_k=\mbf{y}_k-\check{\mbf{y}}_k$ and the corrected covariance is given by \eqref{eq:forward_cov_hat}. \ Note that $\mbsdel{\chi}_{1}^\wedge$ is an element of the Lie algebra associated with $\mc{G}$.

The backward smoothing is initialized using the state estimate and covariance output of the forward filter. The Kalman gain for the smoother is given by \eqref{eq:RTS_smoother_gain} and the smoother covariance update is given by \eqref{eq:RTS_smoother_cov}. The smoother state innovation and smoother state update are given by 
\begin{align}
\begin{bmatrix}
\mbf{z}_{s,k}^1 \\ 
\mbf{z}_{s,k}^2
\end{bmatrix} & = \begin{bmatrix}
{{\log\left(\mbfhat{X}_{s, k+1}^{-1}\mbfch{X}_{f,k+1}\right)}^\vee} \\ 
\mbfhat{x}_{s,k+1}-\check{\mbf{x}}_{f,k+1}
\end{bmatrix}, \\
\mbfhat{X}_{s,k} & = \hat{\mbf{X}}_{f,k}\exp\left(-\left(\mbf{K}_{s,k}\mbf{z}_{s,k}^1\right)^\wedge\right), \\
\mbfhat{x}_{s,k} & = \mbfhat{x}_{f,k} - \mbf{K}_{s,k}\mbf{z}_{s,k}^2.
\end{align}

\section{Invariant RTS Smoothing}
\label{section:IRTS}

The invariant extended Kalman filter (IEKF) \cite{MyPaper:Barrau} can be considered a variant of the extended Kalman filter (EKF), but specific to estimation problems with particular properties. Here, the invariant filtering framework is leveraged to introduce the invariant RTS (IRTS) smoother. Similar to traditional RTS and MRTS smoothing, the forward filter in IRTS smoothing is the IEKF. Subsequently, a backward smoothing step is performed. The IEKF and proposed IRTS smoother are applicable to process models of the form 
\beq
\mbfdot{X}(t) = \mbf{F}\left(\mbf{X}(t),\mbf{u}(t)\right)+\mbf{X}(t)\mbf{W}(t), \label{eq:IRTS_nonlinear_process}
\eeq
where $\mbf{X}(t)\in\mathcal{G}$ is the state and $\mbf{W}(t)\in\mathfrak{g}$ is process noise where $\mbf{w}(t)=\mbf{W}(t)^\vee  \in \mathbb{R}^m$. The multiplication between the state and process noise in \eqref{eq:IRTS_nonlinear_process} is necessary since matrix Lie groups are closed under multiplication but not under addition. The argument $(t)$ will once again be omitted for brevity. Given the true state $\mbf{X}$ and the state estimate $\mbfhat{X}$, the error can be defined as a left-invariant error or right-invariant error. Additionally, when the process model (excluding noise) is group affine, the error dynamics are state-independent, as stated in 
Theorem~\ref{thm:theorem_1}.

\begin{theorem}[State-independent error dynamics \cite{MyPaper:Barrau}]
    If the function $\mbf{F}\left(\mbf{X}(t),\mbf{u}(t)\right)$ is group affine and the error is either left- or right-invariant, then the error propagation will be state independent. \label{thm:theorem_1}
\end{theorem}

The choice of using a left- or right-invariant error depends on the left- or right-invariant form of the measurements, as defined next. 

\noindent \textit{Definition 3 (Left- and right-invariant measurement model):} Left- and right-invariant measurements are 
\begin{align}
\mbf{y}_k^\mathrm{L} & = \mbf{X}_k\mbf{b}_k+\mbf{v}_k, \label{eq:left-invariant_meas} \\
\mbf{y}_k^\mathrm{R} & = \mbf{X}_k^{-1}\mbf{b}_k+\mbf{v}_k, \label{eq:right-invariant_meas}
\end{align}
respectively, where $\mbf{b}_k$ is some known column matrix of appropriate dimension.

When confronted with a left-invariant (right-invariant) measurement model and a group affine process model, a left-invariant error (right-invariant error) should be used \cite{MyPaper:Barrau,MyPaper:Barrau_2}.

The error dynamics are linearized using $\mbfdel{X}=\exp\left(\mbsdel{\xi}^\wedge\right)\approx\mbf{1}+\mbsdel{\xi}^\wedge$, where $\mbsdel{\xi}\in\mathbb{R}^m$ is the state of the linearized system, and the superscript for left or right has been dropped here. The linearized process model is 
\beq
\delta\mbsdot{\xi} = \mbf{A}\mbsdel{\xi}+\mbf{L}\mbfdel{w}. \label{IRTS_CT_process_linear}
\eeq 
The prediction step in the forward filter from $\mbf{X}_{k-1}$ to $\mbf{X}_{k}$ is performed by integrating \eqref{eq:IRTS_nonlinear_process} from time $t_{k-1}$ to $t_{k}$. The predicted covariance is given by \eqref{eq:RTS_cov_check}. 

\subsection{L-IRTS Smoothing}
\label{section:LIRTS}

A detailed explanation of left-IRTS (L-IRTS) smoothing is provided here. Because L-IRTS smoothing and right-IRTS smoothing (R-IRTS) smoothing are required in the sample problem presented in Section~\ref{section:Problem}, the equivalent equations for R-IRTS smoothing are provided in Section~\ref{section:RIRTS}.

The left-invariant error $\mbfdel{X}_{f,k}=\mbfhat{X}_{f,k}^{-1}\check{\mbf{X}}_{f,k}$ is the error between the predicted and the corrected state. Rearranging the error definition results  in an expression for the corrected state given by
\begin{align} \label{eq:liekf_state_correction}
\mbfhat{X}_{f,k} & = \check{\mbf{X}}_{f,k}\mbfdel{X}_{f,k}^{-1}, = \check{\mbf{X}}_{f,k}\exp\left(-\left(\mbf{K}_{f,k}\mbf{z}_{f,k}^{\mathrm{L}}\right)^\wedge\right),
\end{align}
where the left-innovation is given by
    \begin{align} \label{eq:liekf_innovation}
        \mbf{z}_{f,k}^{\mathrm{L}}=\check{\mbf{X}}_{f,k}^{-1}\left(\mbf{y}_k-\check{\mbf{y}}_k\right),
    \end{align}
and the Kalman gain is given by \eqref{eq:Gain_forward}, $\check{\mbf{y}}_k=\check{\mbf{X}}_{f,k}\mbf{b}_{k}$, and the negative sign in the exponent is a result of the inverse operation.  To find the Jacobians $\mbf{H}_k$ and $\mbf{M}_k$ the innovation must be linearized. To do so, substitute \eqref{eq:left-invariant_meas} into the innovation expression and use the definition of the error, which results in 
\beq
    \mbfdel{z}_{f,k}^{\mathrm{L}} = \mbf{H}_k\delta\mbsch{\xi}_{f,k}+\mbf{M}_k\mbfdel{v}_k.
\eeq

After the forward-filter pass, the backward smoothing is initialized with the corrected state estimate $\mbfhat{X}_{f,N}$ and covariance $\mbfhat{P}_{f,N}$. The Kalman gain used in the backward smoothing is given by \eqref{eq:RTS_smoother_gain} and the covariance update is given by \eqref{eq:RTS_smoother_cov}. To find an expression describing the smoothed state estimate, the error between $\hat{\mbf{X}}_{s,k+1}$ and $\check{\mbf{X}}_{f,k+1}$ must be defined. In the traditional RTS smoother there is the $\hat{\mbf{x}}_{s,k+1}-\check{\mbf{x}}_{f,k+1}$ term which can be seen as a $\delta \mbf{x}$ term or as a smoother innovation term. The equivalent error in the present context, when the state is an element of the matrix Lie group $\mc{G}$ and a left-invariant error is employed, is 
\beq 
	\delta\mbf{X}_{s,k}=\hat{\mbf{X}}^{-1}_{s,k+1}\check{\mbf{X}}_{f,k+1},
	\label{eq:IRTS_left_error_smoother} 
\eeq 
where $\exp\left(\delta\mbs{\xi}_{s,k}^\wedge\right)=\delta\mbf{X}_{s,k}$. Using the definition of $\delta \mbs{\xi}_{s,k}$, an expression for the left smoother-state innovation $\mbf{z}_{s,k}^\mathrm{L}$ is
\beq \mbf{z}_{s,k}^{\mathrm{L}}=\mbsdel{\xi}_{s,k}=\log\left(\hat{\mbf{X}}^{-1}_{s,k+1}\check{\mbf{X}}_{f,k+1}\right)^\vee. \label{eq:IRTS_left_innovation} \eeq
Given that the smoother-state innovation has been defined, an expression for the smoother state estimate update can now be found. The left-invariant IRTS smoother update is then
\beq \hat{\mbf{X}}_{s,k}=\hat{\mbf{X}}_{f,k}\exp\left(-\left(\mbf{K}_{s,k}\mbf{z}_{s,k}^{\mathrm{L}}\right)^\wedge\right). \label{eq:IRTS_smoothed_estimate} 
\eeq

\subsection{R-IRTS Smoothing}
\label{section:RIRTS}

The right-invariant error associated with the filter is
\beq 
    \mbfdel{X}_{f,k} = \check{\mbf{X}}_{f,k}\hat{\mbf{X}}^{-1}_{f,k},
\eeq 
which justifies the form of the state correction, 
\beq \label{eq:riekf_state_correction}
    \mbfhat{X}_{f,k} = \exp\left(-\left(\mbf{K}_{f,k}\mbf{z}_{f,k}^{\mathrm{R}}\right)^\wedge\right)\check{\mbf{X}}_{f,k},
\eeq
where the right-innovation is given by 
    \begin{align} \label{eq:riekf_innovation}
        \mbf{z}_{f,k}^{\mathrm{R}} = \mbfch{X}_k(\mbf{y}_k - \mbfch{y}_k) .
    \end{align}
The innovation and state-estimate update used in the smoothing phase are 
\begin{align}
\mbf{z}_{s,k}^{\mathrm{R}} & = \log\left(\check{\mbf{X}}_{f,k+1}\hat{\mbf{X}}^{-1}_{s,k+1}\right)^\vee  , \label{eq:IRTS_right_innovation} \\
	\hat{\mbf{X}}_{s,k} & = \exp\left(-\left(\mbf{K}_k\mbf{z}_{s,k}^{\mathrm{R}}\right)^\wedge\right)\hat{\mbf{X}}_{f,k} . \label{eq:IRTS_smoothed_estimate}
\end{align}

\section{Sample Problem: $SE(3)$ with Bias}
\label{section:Problem}
\subsection{Problem Formulation}
Consider a rigid body free to rotate and translate in three-dimensional space.  Let $\mathcal{F}_a$ be an inertial frame composed of three orthonormal physical basis vectors \cite{MyPaper:Hughes} and $\mathcal{F}_b$ be a frame that rotates with the body. The orientation of $\mathcal{F}_a$ relative to $\mathcal{F}_b$ is described by a direction cosine matrix (DCM) $\mbf{C}_{ab}\in SO(3)$. A physical vector $\underrightarrow{v}$ can be resolved in $\mathcal{F}_a$ as a column matrix $\mbf{v}_a$ or in $\mathcal{F}_b$ as $\mbf{v}_b$, where $\mbf{v}_a=\mbf{C}_{ab}\mbf{v}_b$ and $\mbf{v}_a,\mbf{v}_b\in\mathbb{R}^3$. Point $w$ is a datum point, and point $z$ is fixed to the body. The position of point $z$ relative to point $w$ resolved in $\mathcal{F}_a$ is denoted as $\mbf{r}^{zw}_a \in \mathbb{R}^3$. The velocity of $z$ relative to $w$ with respect to $\mathcal{F}_a$ is denoted as $\dot{\mbf{r}}^{zw}_a=\mbf{v}^{zw/a}_a$ and the angular velocity of $\mathcal{F}_b$ relative to $\mathcal{F}_a$ resolved in $\mathcal{F}_b$ is given by $\mbs{\omega}^{ba}_b\in\mathbb{R}^3$.  

Consider noisy, biased, angular velocity measurements provided by a rate gyro of the form
\begin{align} \label{eq:rate_gyro_measurements}
\mbf{u}_b^1 = \mbs{\omega}_b^{ba}- \mbs{\beta}_b^1 - \mbf{w}_b^1,
\end{align}
where $\mbf{w}_{b}^1 \sim \mathcal{N}\left(\mbf{0},\mbf{Q}^1\right)$, and $\mbs{\beta}_b^1$ is the measurement bias. Additionally, consider noisy, biased, translational velocity measurements given by 
    \begin{align} \label{eq:velocity_measurements}
        \mbf{u}_b^2 = \mbf{C}_{ab}^\trans \mbf{v}_a^{zw/a} - \mbs{\beta}_b^2 - \mbf{w}_b^2,
    \end{align}
where $\mbf{w}_b^2 \sim \mathcal{N} \left(\mbf{0}, \mbf{Q}^2 \right)$, and $\mbs{\beta}_b^2$ is sensor bias. Both biases are random walk processes where 
    $
        \mbsdot{\beta}_b^1 = \mbf{w}_b^3, \; \mbsdot{\beta}_b^2 = \mbf{w}_b^4,
    $
and $\mbf{w}_b^3 \sim \mathcal{N} \left(\mbf{0}, \mbf{Q}^3 \right)$, $\mbf{w}_b^3 \sim \mathcal{N} \left(\mbf{0}, \mbf{Q}^4 \right)$.
The states of interest for estimation are the attitude $\mbf{C}_{ab}$, position $\mbf{r}_a^{zw}$, and sensor biases $\mbs{\beta}_b^1$ and $\mbs{\beta}_b^2$. Similarly to \cite{heo2018consistent}, these states can be cast into an element of a matrix Lie group $\mathcal{G}$ as
	\begin{align}
		\mbf{X} = 
			\begin{bmatrix}  
				\mbf{C}_{ab} & \mbf{r}_a^{zw}\\
				& 1 \\
				& & \mbf{1} & \mbs{\beta}_b^1 & \mbs{\beta}_b^2 \\
				& & & 1 & \\ 
				& & & & 1 
				\end{bmatrix} \in \mathcal{G}.
	\end{align} 
Details pertaining to the matrix Lie group $\mathcal{G}$ are presented in the Appendix. 

The kinematics of the problem are given by  
\begin{align}
\mbfdot{C}_{ab} = \mbf{C}_{ab}{\mbs{\omega}_b^{ba}}^\times, 
 \hspace{5mm} \mbfdot{r}_a^{zw} = \mbf{v}_a^{zw/a}, 
\end{align}
where $\left(\cdot\right)^\times:\mathbb{R}^3 \rightarrow \mathfrak{so}(3)$ is the linear operator that maps a three dimensional column matrix to the matrix Lie algebra $\mathfrak{so}(3)$. The continuous-time process model is then given by
\begin{align}
\mbfdot{C}_{ab} & = \mbf{C}_{ab}\left(\mbf{u}_b^1+ \mbs{\beta}_b^1 + \mbf{w}_b^1\right)^\times, \; \; \; \mbsdot{\beta}_b^1 = \mbf{w}_b^3 , \\
\mbfdot{r}_a^{zw} & = \mbf{C}_{ab}\left(\mbf{u}_b^2+ \mbs{\beta}_b^2 + \mbf{w}_b^2\right) , \; \; \; \mbsdot{\beta}_b^2 = \mbf{w}_b^4 .
\end{align}
The inclusion of sensor biases in the problem violates the group affine properties required by the invariant framework, meaning the kinematics do not satisfy \eqref{eq:group_affine}. However, the use of the invariant error in the linearization leads to Jacobians that are less state-estimate dependent than Jacobians computed using a multiplicative error definition. 

A position measurement resolved in $\mc{F}_a$, such as a global positioning system (GPS) measurement, provides 
\beq \mbf{y}_{a,k} = \mbf{r}_{a}^{z_k w}+\mbf{v}_{a,k} , \label{eq:problem_meas} 
\eeq
where $\mbf{v}_{a,k} \sim \mathcal{N} \left(\mbf{0}, \mbf{R}_{k}^1 \right)$, which can also be written as a function of the state $\mbf{X}_k$, 
\begin{align} \label{eq:gps_measurements}
    \begin{bmatrix} \mbf{y}_{a,k} \\ 1 \\ \mbf{0} \end{bmatrix} = \mbf{X}_k \begin{bmatrix}
    \mbf{0} \\
    1 \\
    \mbf{0}
    \end{bmatrix}+\begin{bmatrix} 
    \mbf{v}_{a,k} \\
    \mbf{0} \\ 
    \mbf{0}
    \end{bmatrix},
\end{align}
where the matrices $\mbf{0}$ are of appropriate size. The position measurement is left-invariant because it is of the form given in \eqref{eq:left-invariant_meas}.
%
In addition, landmark measurements resolved in $\mc{F}_b$, from a LIDAR or stereo camera, for example, are available. Denote point $p^\imath$ to be the $\imath^\mathrm{th}$ landmark. The position of the $\imath^\mathrm{th}$ landmark relative to $w$ resolved in $\mathcal{F}_a$ is given by $\mbf{r}_a^{p_\imath w}$. The landmark sensor measures
    \begin{align}
        \mbf{y}_{b_k}^\imath = \mbf{C}_{ab_k}^\trans \left(\mbf{r}_a^{p_\imath w} - \mbf{r}_a^{z_k w} \right) + \mbf{v}_{b_k}^\imath,
    \end{align}
where $\mbf{v}_{b_k}^2 \sim \mathcal{N} \left(\mbf{0}, \mbf{R}_k^\imath \right)$, which can alternatively be written
    \begin{align} \label{eq:camera_measurements}
        \begin{bmatrix} \mbf{y}_{b_k}^\imath \\ 1 \\ \mbf{0} \end{bmatrix} =  \mbf{X}_k^{-1} \begin{bmatrix} \mbf{r}_a^{p_\imath w} \\ 1 \\ \mbf{0} \end{bmatrix} + \begin{bmatrix} \mbf{v}_{b_k}^\imath \\ 0 \\ \mbf{0} \end{bmatrix}
    \end{align}
The landmark sensor measurement is right-invariant because it is of the form given in \eqref{eq:right-invariant_meas}.

\subsection{Using both Left- and Right-Invariant Measurements} \label{sec:ert}

%
%
Generally, the choice of the left- or right-invariant error is dictated by the left- or right-invariant form of the exteroceptive measurements, as described in Section~\ref{section:IRTS}.
For instance, given only a left-invariant measurement, such as GPS, the L-IRTS smoothing approach of Section~\ref{section:LIRTS} would be employed. However, when both left- and right-invariant measurements are available, such as in this paper, in order to maintain innovation Jacobians in the forward pass that are state-estimate independent, an error representation transformation (ERT) between left- and right-invariant errors is done, as discussed in \cite{hartley2019contact}. Nominally, a left-invariant correction, given by~\eqref{eq:liekf_state_correction}, and covariance computation, given by~\eqref{eq:forward_cov_hat}, are computed in the forward pass of the smoother. When a right-invariant measurement is available, the left-invariant covariance, $\mbfcheck{P}_k^\textrm{L}$, is mapped to the right-invariant covariance, $\mbfcheck{P}_k^\textrm{R}$, using the matrix representation of the adjoint operator, 
    \begin{align} \label{eq:switch}
        \mbfcheck{P}_k^\textrm{R} = \mathrm{Ad} \left(\mbfcheck{X}_k \right) \mbfcheck{P}_k^\textrm{L} \mathrm{Ad} \left(\mbfcheck{X}_k \right)^\trans.
    \end{align}
The right-invariant measurement is then used within a right-invariant correction step, given by~\eqref{eq:riekf_state_correction}, and covariance correction, again using~\eqref{eq:forward_cov_hat} but with the appropriate right-invariant Jacobians. The corrected covariance in right-invariant form is then mapped back to a covariance in left-invariant form using an analogous version of~\eqref{eq:switch}. This ERT of the covariance between left-and right-invariant forms ensures that the Jacobians used within the correction step and covariance computation~\eqref{eq:forward_cov_hat} are state-estimate independent. 

When solving the batch state estimation problem with both left- and right-invariant measurement using IGN the measurements are not processed sequentially. Rather, all measurements, both left- and right-invariant, are used simultaneously. Therefore, when using a left-invariant error and innovation, although the Jacobians associated with the left-invariant measurement will be state-estimate independent, the Jacobians associated with the right-invariant measurement will depend on the state estimate. As such, an advantage of the IRTS smoother over IGN is that the ERT allows for computation of Jacobians that are consistent with the left- or right-invariant form of the measurement.

\subsection{Error Propagation}
To derive the error propagation in the IRTS, a left-invariant error of the form $\delta \mbf{X} = \mbf{X}^{-1} \mbfhat{X}$ is chosen. For this particular problem, the use of a left-invariant error leads to process model Jacobians that are less state-estimate dependent than when a right-invariant error is used. 

Linearizing the process model using the left-invariant error definition leads to 
    \begin{align}
        \delta \mbsdot{\xi} = \mbf{A} \delta \mbs{\xi} + \mbf{L} \delta \mbf{w},
    \end{align}
where 
	\begin{align} \label{eq:irts_linearization}
		\mbf{A} = \begin{bmatrix} -(\mbf{u}_b^1 + \mbshat{\beta}_b^1)^\times & \mbf{0} & \mbf{1} & \mbf{0} \\ -\left(\mbf{u}_b^2 + \mbshat{\beta}_b^2 \right)^\times & -\left(\mbf{u}_b^1 + \mbshat{\beta}_b^1 \right)^\times & \mbf{0} & \mbf{1} \\ \mbf{0} & \mbf{0} & \mbf{0} & \mbf{0}  \\ \mbf{0} & \mbf{0} & \mbf{0} & \mbf{0}  \end{bmatrix}, \hspace{2mm} \mbf{L} = - \mbf{1},
	\end{align}
and $\delta \mbs{\xi} = \begin{bmatrix} \delta \mbs{\xi}^{\phi^\trans} & \delta \mbs{\xi}^{\mathrm{r}^\trans} & \delta \mbs{\xi}^{\beta^{1^\trans}} & \delta \mbs{\xi}^{\beta^{2^\trans}}  \end{bmatrix}^\trans$.
Notice that $\mbf{A}$ depends on the measurements $\mbf{u}_b^1$ and $\mbf{u}_b^2$, the bias estimates $\mbshat{\beta}_b^1$ and $\mbshat{\beta}_b^2$, but not on any other state estimates. 

The continuous-time process model can then be discretized using any desired discretization scheme. In this paper, a forward Euler discretization scheme is chosen for simplicity.


\subsection{Linearization of the Measurement Model}

For the left-invariant measurements the innovation term is given by~\eqref{eq:liekf_innovation}, and for the right-invariant measurements the innovation term is given by~\eqref{eq:riekf_innovation}. To find the Jacobians associated with the position measurements, the innovation $\mbf{z}_{f,k}^\mathrm{L}$ is linearized using the left-invariant error definition. Similarly, to find the Jacobians associated with the landmark measurements, the innovation $\mbf{z}_{f,k}^\mathrm{R}$ is linearized using the right-invariant error definition. Linearizing the innovation $\mbf{z}_{f,k}^\mathrm{L}$ leads to measurement model Jacobians of the form
    \begin{align}
        \mbf{H}_k^\mathrm{L} = \begin{bmatrix} \mbf{0} & \mbf{-1} & \mbf{0} & \mbf{0} \end{bmatrix}, \hspace{5mm} \mbf{M}_k^\mathrm{L} = \mbfcheck{C}_{ab_k}^\trans.
    \end{align}
For each of the $m$ landmarks, linearizing the innovation $\mbf{z}_{f,k}^\mathrm{R}$ using the right-invariant error definition leads to measurement model Jacobians of the form 
    \begin{align}
        \mbf{H}_k^{\mathrm{R}, \imath} & = \underset{\imath = 1, \ldots, m}{\mathrm{row}} \begin{bmatrix} -\mbf{r}_a^{p_\imath w^\times} & \mbf{1} & \mbf{0} & \mbf{0} \end{bmatrix}, \label{eq:R_inv_cam_meas}\\
        \mbf{M}_k^\mathrm{R} & = \mathrm{diag} \left(\mbfcheck{C}_{ab_k}, \ldots, \mbfcheck{C}_{ab_k} \right).
    \end{align}
    
Note that the Jacobians $\mbf{H}_k^\mathrm{L}$ and $\mbf{H}_k^{\mathrm{R}, \imath}$ are constant and do not depend on the state, as the position of the landmarks $\mbf{r}_a^{p_\imath w}$ are known.

\subsection{Jacobians using MRTS smoothing}
To implement the MRTS smoother, attitude error is taken to be multiplicative, while position and bias errors are taken to be additive. The error definitions for the MRTS smoother used in the linearization of the process and measurement model are given by 
	\begin{align} \label{eq:mrts_error_defintions}
		\delta \mbf{C} & = \mbf{C}_{ab}^\trans \mbfhat{C}_{ab}, \hspace{5mm}
		\delta \mbf{r}_a^{zw} = \mbfhat{r}_a^{zw} - \mbf{r}_a^{zw}, \\
		\delta \mbs{\beta}_b^1 & = \mbshat{\beta}_b^1 - \mbs{\beta}_b^1, \hspace{5mm}
		\delta \mbs{\beta}_b^2 = \mbshat{\beta}_b^2 - \mbs{\beta}_b^2.
	\end{align}
    
The Jacobians associated with MRTS smoother, denoted by the $\left( \cdot \right)^\star$, are given by
	\begin{align}
		\mbf{A}^\star & = \begin{bmatrix} -(\mbf{u}_b^1 + \mbshat{\beta}_b^1)^\times & \mbf{0} & \mbf{1} & \mbf{0} \\ -\mbfhat{C}_{ab} \left(\mbf{u}_b^2 + \mbshat{\beta}_b^2 \right)^\times & \mbf{0} & \mbf{0} & \mbfhat{C}_{ab} \\ \mbf{0} & \mbf{0} & \mbf{0} & \mbf{0} \\ \mbf{0} & \mbf{0} & \mbf{0} & \mbf{0}  \end{bmatrix}, \label{eq:mrts_A} \\
		\mbf{L}^\star & = -\mathrm{diag} \left(\mbf{1}, \mbfhat{C}_{ab}, \mbf{1}, \mbf{1} \right). \label{eq:mrts_L}
	\end{align}
The measurement model Jacobians for the MRTS smoother are derived by linearizing the innovation $\mbf{z}_{f,k} = \mbf{y}_k - \mbfcheck{y}_k$ using the error definitions given by~\eqref{eq:mrts_error_defintions}. The position measurement Jacobians are given by
    \begin{align}
        \mbf{H}_k^{1^\star} = \begin{bmatrix} \mbf{0} & -\mbf{1} & \mbf{0} & \mbf{0} \end{bmatrix}, \hspace{5mm} \mbf{M}_k^{1^\star} = \mbf{1}.
    \end{align}
The Jacobians corresponding to the landmark sensor measurements for the MRTS smoother are given by
    \begin{align}
        \mbf{H}_k^{2, \imath^\star} & = \underset{\imath = 1, \ldots, m}{\mathrm{row}} \begin{bmatrix} - \left(\mbfcheck{C}_{ab_k} \left(\mbf{r}_a^{p_\imath w} - \mbfcheck{r}_a^{z_kw} \right) \right)^\times & \mbfcheck{C}_{ab_k}^\trans & \mbf{0} & \mbf{0} \end{bmatrix}, \label{eq:mrts_H2} \\
        \mbf{M}_k^{2^\star} & = \mbf{1}.
    \end{align}

Notice that the process model Jacobians derived using the error definitions for the MRTS smoother, \eqref{eq:mrts_A} and \eqref{eq:mrts_L}, depend on the attitude estimate $\mbfhat{C}_{ab}$, and not just the bias estimates like the Jacobians used in the IRTS smoother in~\eqref{eq:irts_linearization}. In addition, the measurement model Jacobian $\mbf{H}_k^{2, \imath^\star}$ given in \eqref{eq:mrts_H2}  depends on the attitude estimate and the position estimate, while the Jacobian for the IRTS smoother, $\mbf{H}_k^{\mathrm{R}, \imath}$ given in \eqref{eq:R_inv_cam_meas}, is constant.  



\section{Experimental Results Using the Starry Night Dataset}
\label{section:experimental_results}
The MRTS and the IRTS smoothers were first tested using simulated data, demonstrating encouraging results. To test the smoothers on a real-life problem, the IRTS and MRTS smoothers were compared using the Starry Night dataset \cite{starryNightDataset}. The Starry Night dataset provides angular and translational velocity sensor measurements at approximately \SI{100}{\hertz}, and stereo image pairs logged at approximately \SI{15}{\hertz}. The camera measurements are preprocessed as in \cite{Arsenault2019} to obtain a right-invariant measurement model of the form~\eqref{eq:camera_measurements}. These preprocessed camera measurements are used in both the IRTS and MRTS smoothers. The angular and translational velocity sensor data provided in the dataset are unbiased. To add complexity to the experiments, large artificial biases are added to the angular and translational velocity sensor data. The initial biases are set to $\mbs{\beta}_{b,0}^1 = \begin{bmatrix} 0.05 & 0.05 & 0.05 \end{bmatrix}^\trans$ \SI{}{\radian\per\second} and $\mbs{\beta}_{b,0}^2 = \begin{bmatrix} 0.04 & -0.03 & 0.06 \end{bmatrix}^\trans$ \SI{}{\meter\per\second}. For all experiments, the standard deviation on the bias random walks are both set to $\sigma_{\beta_1} = \sigma_{\beta_2} = 0.005$ with appropriate units. In addition, artificial position measurements of the form~\eqref{eq:gps_measurements} are generated using ground truth data at approximately \SI{10}{\hertz}. The same standard deviation of $\sigma_{R_1} =$ \SI{0.5}{\meter} is used for the artificial position measurements in all subsequent experiments.

\begin{figure}[h] 
	\centering
	\captionsetup{justification=centering}
        		\includegraphics[width=0.45\textwidth]{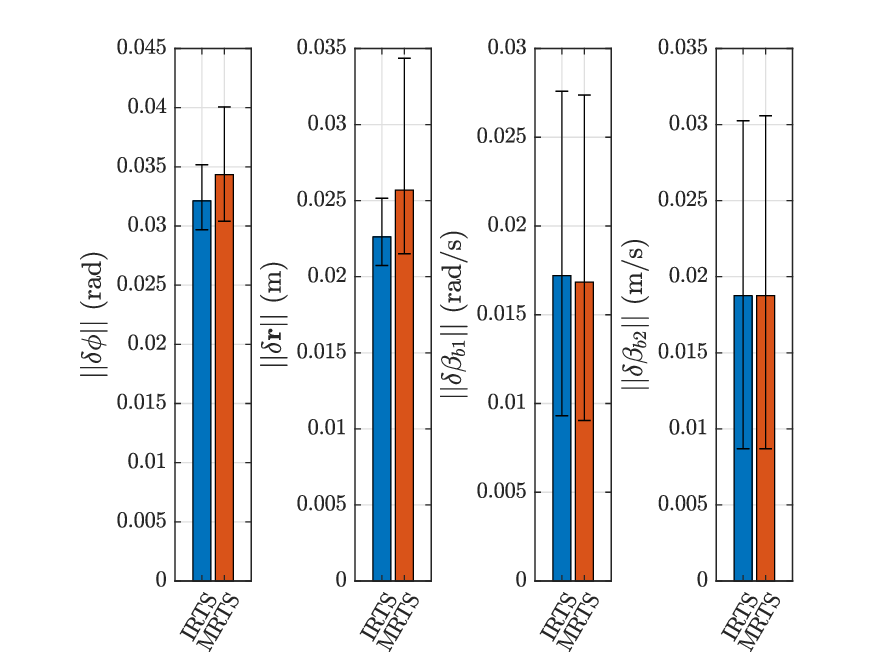}
     	\caption{Mean RMSEs for the IRTS and MRTS smoothers on experimental data with low initialization error.}
    	\label{fig:low_initial_error}
\end{figure}

A $20$ second window of the Starry Night dataset is chosen to test the smoothers. The smoothers are tested for both low and high initialization errors. Figure~\ref{fig:low_initial_error} shows the root mean square error (RMSE) in each state for 100 Monte-Carlo simulations using the IRTS and the MRTS smoothers with a low initialization error. The initial error for each Monte-Carlo simulation is sampled from the distribution $\delta \mbs{\xi}_0 \sim \mathcal{N} \left(\mbf{m}, \mbf{P} \right)$, with the initial mean error set to $\mbf{m} =  
        \bbm
            m_\phi \mbf{1}_{1:3}^\trans &  m_\mathrm{r} \mbf{1}_{1:3}^\trans &  m_{\beta_1} \mbf{1}_{1:3}^\trans &  m_{\beta_2} \mbf{1}_{1:3}^\trans 
        \ebm^\trans,
$
where $\mbf{1}_{1:3} = [ 1 \; 1 \; 1 ]^\trans$, $m_\phi = \frac{\pi}{12}$ \SI{}{\radian}, $m_\mathrm{r} = $ \SI{0.1}{\meter}, $m_{\beta_1} = $ \SI{0.005}{\radian\per\second}, and $m_{\beta_2} = $ \SI{0.005}{\meter\per\second}. The covariance on the initial error distribution is set to $\mbf{P} = \mathrm{diag} \left(\sigma_\phi^2 \mbf{1}, \sigma_\mathrm{r}^2 \mbf{1}, \sigma_{\beta_1^0}^2 \mbf{1}, \sigma_{\beta_2^0}^2 \mbf{1} \right),$     
where $\sigma_\phi = \frac{\pi}{36} \hspace{1mm} \mathrm{rad}$, $\sigma_\mathrm{r} = 0.1 \hspace{2mm} \mathrm{m}$, $\sigma_{\beta_1^0} =$ \SI{0.005}{\radian\per\second}, and $\sigma_{\beta_2^0} =$ \SI{0.005}{\meter\per\second}. In addition, for each Monte-Carlo trial, new realizations of the artificial biases and artificial position measurements are generated. 
The initial state estimate for the IRTS smoother is generated using the left-invariant error definition, while the initial state estimate for the MRTS smoother is generated using the multiplicative error definitions given by \eqref{eq:mrts_error_defintions}.
The lower bounds of the error bars in Figure \ref{fig:low_initial_error}, and all forthcoming figures, are set to a percentile of 2.5 and the upper bounds of the error bars are set to a percentile of 97.5, meaning the results of 95\% of the trials lie within the error bars.

\begin{figure}[h]
	\centering
	\captionsetup{justification=centering}
        		\includegraphics[width=0.45\textwidth]{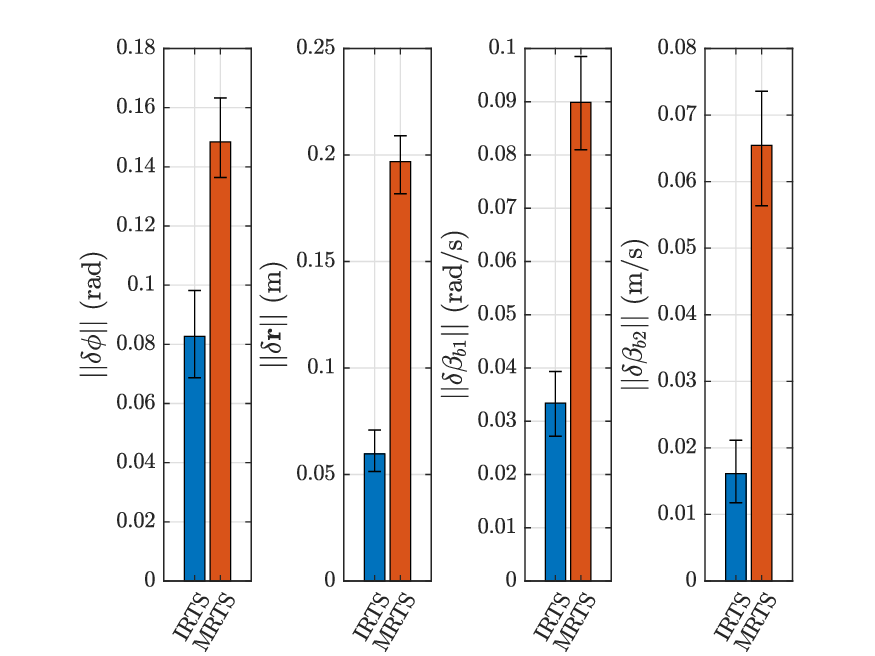}
     	\caption{Mean RMSEs for the IRTS and MRTS smoothers on experimental data with high initialization error.}
    	\label{fig:high_initial_error}
\end{figure}

With a low initialization error, Figure~\ref{fig:low_initial_error} indicates that the performance of the IRTS and MRTS smoothers are comparable, with the IRTS smoother performing slightly better on average. The advantage of the IRTS smoother lies in the case of poor initialization, as the Jacobians of the IRTS smoother are less state-estimate dependent and are therefore more accurate even when the state estimate is poor. To demonstrate this, both filters were tested for a large initialization error in all states. The mean initial errors for each state are set to  $m_\phi = \frac{\pi}{3}$ \SI{}{\radian}, $m_\mathrm{r} =$ \SI{1}{\meter}, $m_{\beta_1} =$ \SI{0.03}{\radian\per\second}, and $m_{\beta_2} =$ \SI{0.03}{\meter\per\second}. Figure~\ref{fig:high_initial_error} presents the results of 100 Monte-Carlo simulations. Figure~\ref{fig:high_initial_error} demonstrates that when the smoothers are poorly initialized, the IRTS smoother outperforms the MRTS smoother by a large margin.

The IRTS and MRTS smoothers are batch state estimators, which in turn motivates a comparison to IGN and MGN, for one and more iterations. The nonlinear least squares problem associated with a \emph{maximum a posteriori} (MAP) formulation of the state estimation problem is solved via GN optimization \cite{MyPaper:barfoot}. Within the IGN framework, a left-invariant error and innovation are employed with left-invariant measurements, while a left-invariant error and standard innovation is used with right-invariant measurements. Within the MGN framework, the multiplicative errors defined in~\eqref{eq:mrts_error_defintions} and a standard innovation are used for all measurements. 



\begin{figure}[h]
	\centering
	\captionsetup{justification=centering}
        		\includegraphics[width=0.5\textwidth]{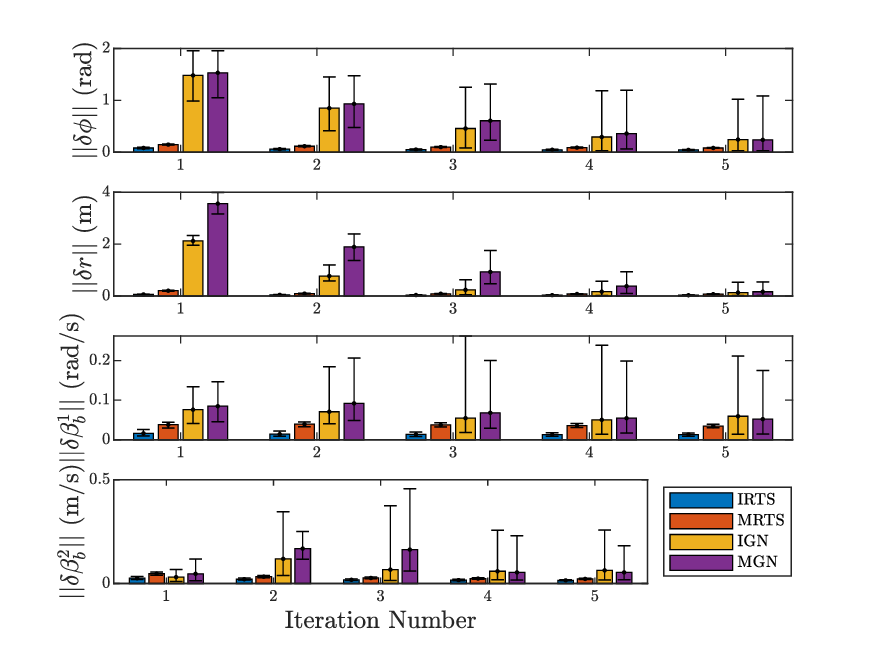}
     	\caption{Mean RMSEs for each smoother and Gauss-Newton algorithms for each iteration.}
    	\label{fig:RMSE_barfoot_GN}
\end{figure}

Next, a comparison of the IRTS and MRTS smoothers to both IGN and MGN is provided.  
One iteration of either the IRTS or the MRTS smoothers is defined as one forward pass followed by one backwards pass of the smoothing algorithm. One iteration of a smoothing algorithm is almost, but not exactly, equivalent to one iteration of GN \cite{Aravkin2014}. For every subsequent iteration of the smoothers, the forward filter is then re-initialized with the new initial state and covariance estimates from the previous backwards pass. Both IGN and MGN are initialized using dead reckoning. The mean RMSEs after each iteration of the smoothers and GN estimators are compared over 100 Monte-Carlo trials in Figure~\ref{fig:RMSE_barfoot_GN}. In each Monte-Carlo simulation, a high initialization error is tested with the same parameters used previously. 

The results in Figure~\ref{fig:RMSE_barfoot_GN} demonstrate that one iteration of either smoothers far outperforms one iteration of either of the GN algorithms. The reason is that the Jacobians in the smoothers are computed using the best available state estimate in both the forward and backwards passes, meaning the Jacobians used in the smoothers are more accurate as the forward and backward passes are executed step by step. In both IGN and MGN, all Jacobians are evaluated using the state estimate from the previous least-squares solution.T The Jacobians in the first iteration of each GN a   lgorithm are inaccurate due to their computation using dead-reckoning starting from a large initial error. Even after 5 iterations of both IGN and MGN, there is still large variability in the mean RMSEs, as shown by the top of the error bars in Figure~\ref{fig:RMSE_barfoot_GN}. Both of the smoothers show significantly smaller mean RMSEs after one iteration, but also much lower variability in the results. It was noted that using a lower initialization error leads to a less drastic difference between one iteration of the smoothers and one iteration of GN approaches because the Jacobians used in the first iteration of the GN approaches are more accurate, and the solution converges faster.




There are further notable advantages to the IRTS smoother over IGN. For example, the covariance computation is straightforward in the IRTS smoother. In the forward pass, the covariance is computed using \eqref{eq:forward_cov_hat}, and in the backward pass, the covariance is computed using \eqref{eq:RTS_smoother_cov}. On the other hand, extracting the covariance associated with each state at each time step is cumbersome in IGN and MGN because a large, sparse, matrix must be inverted. Additionally, in a problem with both left- and right-invariant measurements, such as the problem presented in this paper, the ERT can be used, as discussed in Section~\ref{sec:ert}. This leads to a correction step in the forward pass that is always consistent with the measurement type (i.e., consistent left- or right-invariant measurements). In an IGN framework, the innovation is fixed and cannot be changed, ultimately leading to Jacobians in the IGN framework that are state-estimate dependent in a problem with both left- and right-invariant measurements.


\section{Conclusions}
\label{section:Conclusions}

%
The main purpose of this paper is to present the IRTS smoother and benchmark it using experimental data relative to the MRTS smoother, IGN, and MGN. 
The IRTS smoother is essentially an application of the invariant filtering of \cite{MyPaper:Barrau, MyPaper:Barrau_2} where left- or right-invariant error definitions, group-affine process models, and left- or right-invariant measurement models are leveraged to give state-independent Jacobians. The IRTS and MRTS smoother were compared on an $SE(3)$ problem where the attitude and position of a body, and sensor biases, are estimated using angular velocity data, translational velocity data, stereo camera data, and GPS measurements simulated from ground-truth data. The IRTS smoother performance is quite good, performing as well or better than all other algorithms. Specifically, when the initialization is poor, the IRTS smoother provides significantly better performance than the MRTS smoother. Furthermore, several iterations of IGN and MGN were required to match the performance of the IRTS smoother.


%

\appendices
\section{$SE(3)$ with Bias}
Three dimensional pose and two sensor biases can be cast into a matrix Lie group $\mathcal{G}$ of the form 
	\begin{align}
		\mbf{X} = 
			\begin{bmatrix}  
				\mbf{C} & \mbf{r}\\
				& 1 \\
				& & \mbf{1} & \mbs{\beta}_b^1 & \mbs{\beta}_b^2 \\
				& & & 1 & \\ 
				& & & & 1 
				\end{bmatrix} \in \mathcal{G},
	\end{align} 
where $\mathcal{G} \in \mathbb{R}^{9 \times 9}$, $\mbf{C} \in SO(3)$, and $\mbf{r}, \mbs{\beta}_b^1, \mbs{\beta}_b^2 \in \mathbb{R}^3$, and zero entries have been omitted.

The matrix Lie algebra associated with $\mathcal{G}$ is
\begin{align}
	\mathfrak{g} = \left\{ \mbs{\Xi} = \mbs{\xi}^\wedge \in \mathbb{R}^{12 \times 12} \ | \ \mbs{\xi} \in \mathbb{R}^{12}\right\},
\end{align}
where 
\begin{align}
    \mbs{\xi}^\wedge = 
        \begin{bmatrix}
            \mbs{\xi}^\phi \\
            \mbs{\xi}^\mathrm{r} \\
            \mbs{\xi}^{\mbs{\beta}^1} \\
            \mbs{\xi}^{\mbs{\beta}^2}
        \end{bmatrix} = 
        \begin{bmatrix} 
            {\mbs{\xi}^{\phi}}^{\times} & \mbs{\xi}^\mathrm{r} \\
            & 0 \\
            & & \mbf{0} & \mbs{\xi}^{\mbs{\beta}^1} & \mbs{\xi}^{\mbs{\beta}^2} \\
            & & & 0 & \\
            & & & & 0
        \end{bmatrix}.
\end{align}
Additionally, $\mathfrak{so}(3) = \left\{ \mbs{\Omega} = \mbs{\omega}^\times \in \mathbb{R}^{3 \times 3} \ | \ \mbs{\omega} \in \mathbb{R}^3 \right\}$ where
$$
	\mbs{\omega}^\times
	=
	\bma{ccc}
		0 & -\omega_3 & \omega_2 \\
		\omega_3 & 0 & -\omega_1 \\
		-\omega_2 & \omega_1 & 0
	\ema .
$$
The exponential map from $\mathfrak{g}$ to $\mathcal{G}$ is 
\begin{align}
	\exp\left(\mbs{\xi}^\wedge\right) = 
	\begin{bmatrix}
		\exp_{SO(3)}\left({\mbs{\xi}^\phi}^\times\right) & \mbf{J}\mbs{\xi}^\mathrm{r} \\
		 & 1 \\
		 & & \mbf{1} &  \mbs{\xi}^{\mbs{\beta}^1} & \mbs{\xi}^{\mbs{\beta}^2} \\
		 & & & 1 \\
		 & & & & 1
	\end{bmatrix} \in \mathcal{G}.
\end{align}
where $\exp_{SO(3)}\left({\mbs{\xi}^\phi}^\times\right)$ is given by Rodrigues' rotation formula \cite{MyPaper:barfoot} and $\mbf{J}$ is given by 
\beq
\mbf{J} = \f{\sin\phi}{\phi}\mbf{1}+\left(1-\f{\sin\phi}{\phi}\right)\mbf{a}\mbf{a}^\trans+\f{1-\cos\phi}{\phi}\mbf{a}^\times, \nonumber
\eeq
where $\phi=\|\mbs{\xi}^\phi\|$ and $\mbf{a}=\mbs{\xi}^\phi/\phi$. The logarithmic map from $\mathcal{G}$ to $\mathfrak{g}$ is given by
    \begin{align}
        \mathrm{log} \left(\mbf{X} \right) =
            \begin{bmatrix}
                \mathrm{log}_{SO(3)} \left(\mbf{C} \right) & \mbf{J}^{-1} \mbf{r} \\
                & 0 \\
                & & 0 & \mbs{\beta}_b^1 & \mbs{\beta}_b^2 \\
                & & & 0 \\
                & & & & 0
            \end{bmatrix} \in \mathfrak{g}.
    \end{align}
The adjoint operator for the group $\mathcal{G}$ is given by
    \begin{align}
        \mathrm{Ad} \left(\mbf{X} \right) = \begin{bmatrix} \mbf{C} \\
        \mbf{r}^\times \mbf{C} & \mbf{C} \\
        & & \mbf{1} \\
        & & & \mbf{1} 
        \end{bmatrix}.
    \end{align}

\section*{Acknowledgment}

The authors graciously acknowledge funding from Group for Research in Decision Analysis (GERAD) and the National Science and Engineering Research Council (NSERC) of Canada. 

\ifCLASSOPTIONcaptionsoff
  \newpage
\fi



%
\bibliographystyle{IEEEtran}
\bibliography{MyPaper}

%




\end{document}